Data and text mining

# Graph embedding on biomedical networks: methods, applications and evaluations

**Xiang Yue[1,*], Zhen Wang[1], Jingong Huang[2], Srinivasan Parthasarathy[1], Soheil Moosavinasab[3], Yungui Huang[3], Simon M. Lin[3], Wen Zhang[4], Ping Zhang[1,5] and Huan Sun[1,*]**

[1]Department of Computer Science and Engineering, [2]Department of Electrical and Computer Engineering, The Ohio State University, Columbus, OH, USA, [3]Research Information Solutions and Innovation, The Research Institute at Nationwide Children's Hospital, Columbus, OH, USA, [4]College of Informatics, Huazhong Agricultural University, Wuhan, Hubei, China and [5]Department of Biomedical Informatics, The Ohio State University, Columbus, OH, USA

*To whom correspondence should be addressed.

## Abstract

**Motivation:** Graph embedding learning that aims to automatically learn low-dimensional node representations, has drawn increasing attention in recent years. To date, most recent graph embedding methods are evaluated on social and information networks and are not comprehensively studied on biomedical networks under systematic experiments and analyses. On the other hand, for a variety of biomedical network analysis tasks, traditional techniques such as matrix factorization (which can be seen as a type of graph embedding methods) have shown promising results, and hence there is a need to systematically evaluate the more recent graph embedding methods (e.g. random walk-based and neural network-based) in terms of their usability and potential to further the state-of-the-art.

**Results:** We select 11 representative graph embedding methods and conduct a systematic comparison on 3 important biomedical *link prediction* tasks: drug-disease association (DDA) prediction, drug–drug interaction (DDI) prediction, protein–protein interaction (PPI) prediction; and 2 *node classification* tasks: medical term semantic type classification, protein function prediction. Our experimental results demonstrate that the recent graph embedding methods achieve promising results and deserve more attention in the future biomedical graph analysis. Compared with three state-of-the-art methods for DDAs, DDIs and protein function predictions, the recent graph embedding methods achieve competitive performance without using any biological features and the learned embeddings can be treated as complementary representations for the biological features. By summarizing the experimental results, we provide general guidelines for properly selecting graph embedding methods and setting their hyper-parameters for different biomedical tasks.

**Availability and implementation:** As part of our contributions in the paper, we develop an easy-to-use Python package with detailed instructions, BioNEV, available at: https://github.com/xiangyue9607/BioNEV, including all source code and datasets, to facilitate studying various graph embedding methods on biomedical tasks.

**Contact:** yue.149@osu.edu or sun.397@osu.edu

**Supplementary information:** Supplementary data are available at *Bioinformatics* online.

## 1 Introduction

Graphs (a.k.a. networks) have been widely used to represent biomedical entities (as nodes) and their relations (as edges). Analyzing biomedical graphs can greatly benefit various important biomedical tasks, such as predicting potential drug indications (a.k.a. drug repositioning) based on drug-disease association (DDA) graphs (Gottlieb *et al.*, 2011), detecting long non-coding RNA (lncRNA) functions based on lncRNA–protein interaction networks (Zhang *et al.*, 2018d), and assisting clinical decision making via disease-symptom graphs (Rotmensch *et al.*, 2017).

In order to analyze the graph data, a surge of graph embedding (a.k.a. network embedding or graph representation learning) methods (Grover and Leskovec, 2016; Perozzi *et al.*, 2014; Ribeiro *et al.*, 2017; Tang *et al.*, 2015) have been proposed, where their goal is to automatically learn a low-dimensional feature representation for each node in the graph. Intuitively, the low-dimensional representations are learned to preserve the structural information of graphs, and thus can be used as features in building machine learning models for various downstream tasks, such as link prediction, community detection, node classification and clustering. However, to date, these advanced approaches are mainly evaluated on non-biomedical networks such as social networks, citation networks, and user-item networks, and only a few studies provide evaluations and analyses on biomedical networks. For example, Nelson *et al.* (2019) review the application of embedding methods on three representative biomedical prediction tasks. For each task, they select two biomedical embedding methods for comparison. But some of the selected methods are biomedical task-driven and may not be generalized to other tasks. And there exist more graph embedding methods in open-domain that need comprehensive comparison. Some recent studies (Hamilton *et al.*, 2017; Su *et al.*, 2018; Zhang *et al.*, 2018a) review the technical details of graph embedding methods, but few of them have systematically compared the performance of each method on biomedical datasets.

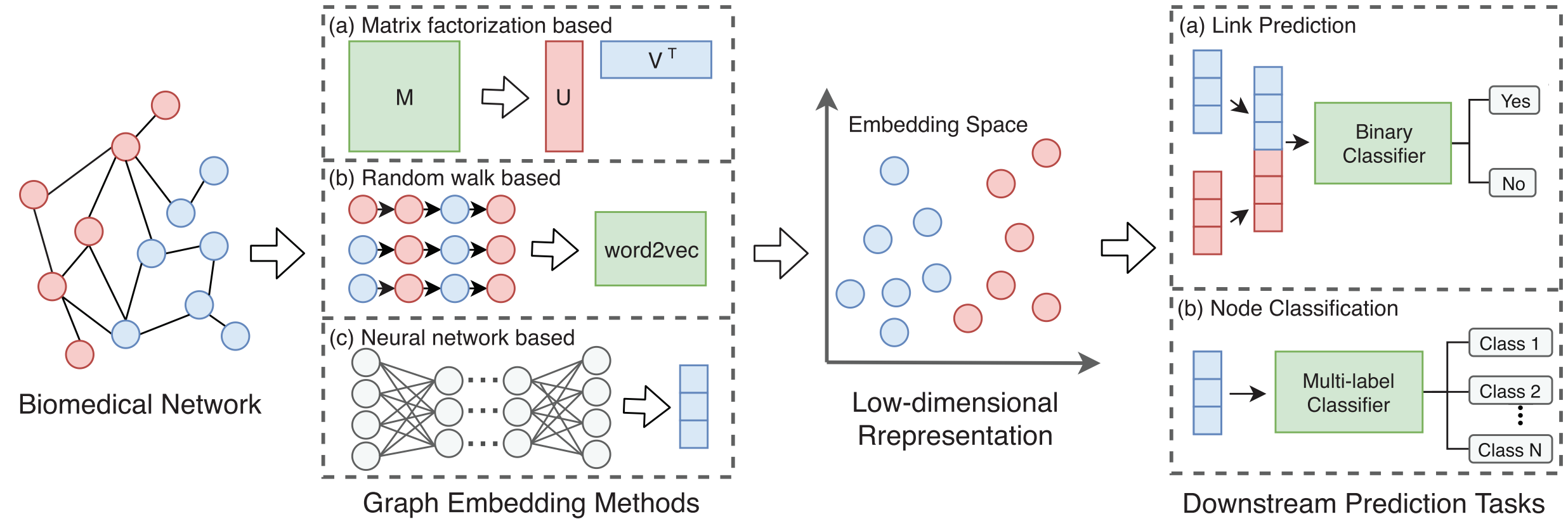

**Fig. 1.** Pipeline for applying graph embedding methods to biomedical tasks. Low-dimensional node representations are first learned from biomedical networks by graph embedding methods and then used as features to build specific classifiers for different tasks. For (**a**) matrix factorization-based methods, they use a data matrix (e.g. adjacency matrix) as the input to learn embeddings through matrix factorization. For (**b**) random walk-based methods, they first generate sequences of nodes through random walks and then feed the sequences into the word2vec model (Mikolov *et al.*, 2013) to learn node representations. For (**c**) neural network-based methods, their architectures and inputs vary from different models (see Section 2 for details)

On the other hand, traditional embedding techniques such as Laplacian eigenmap (LE) (Belkin and Niyogi, 2003) and matrix factorization (MF) have shown promising results for a variety biomedical graph analysis tasks (Ezzat *et al.*, 2017; You *et al.*, 2017). Given that the recent graph embedding methods have been demonstrated to be more effective than the traditional methods in a wide range of non-biomedical tasks (Grover and Leskovec, 2016; Perozzi *et al.*, 2014; Tang *et al.*, 2015), we conduct this work to *investigate the effectiveness and potential of advanced graph embedding methods on biomedical tasks*. Figure 1 summarizes the pipeline for applying various graph embedding methods to downstream prediction tasks.

In this paper, we first provide an overview of existing graph embedding methods and their applications on three important biomedical *link prediction* tasks: DDA prediction (Gottlieb *et al.*, 2011), drug–drug interaction (DDI) prediction (Zhang *et al.*, 2018b), protein–protein interaction (PPI) prediction (Wang *et al.*, 2017) and one popular node classification task, protein function prediction (Cho *et al.*, 2016). In addition, we formulate a relatively less-studied but meaningful node classification task, medical term semantic type classification and apply graph embedding methods to solve it. For the above 5 tasks, we compile 7 datasets from commonly used biomedical databases or previous studies and select 11 graph embedding methods (including both traditional and more recent methods) for comprehensive comparisons. By benchmarking them, we demonstrate that the recent graph embedding methods can achieve promising results in various biomedical tasks and should deserve more attention in the future biomedical graph analysis. Additionally, we compare the graph embedding methods with three recent computational methods that are specially designed for DDAs, DDIs and protein function prediction.

The results indicate that the graph embedding methods can achieve very competitive or better performance while being general (i.e. applied on different graphs and tasks). The learned embedding can also be treated as a complementary representation for the biological features. By summarizing the experimental results, we provide insightful observations as well as suggestions for selecting proper graph embedding methods and setting their hyper-parameters for biomedical prediction tasks. For instance, for MF-based methods, modeling high-order proximity (Cowen *et al.*, 2017) [e.g. HOPE (Ou *et al.*, 2016) and GraRep (Cao *et al.*, 2015)] is more useful in biomedical link prediction tasks compared with node classification tasks. For random walk-based methods, DeepWalk (Perozzi *et al.*, 2014) and node2vec (Grover and Leskovec, 2016) perform better in node classification tasks while struc2vec achieves better results in biomedical link prediction tasks. We also discuss the connections between embedding methods and the recent network propagation and diffusion methods in biomedical graph analysis (Cowen *et al.*, 2017). Additionally, we illustrate a few new trends and directions (e.g. transfer learning in biomedical graph embedding) to encourage future work.

To summarize, our contributions are 3-fold:

- We provide an overview of different types of graph embedding methods, and discuss how they can be used in three important biomedical *link prediction* tasks: DDAs, DDIs and PPIs prediction; and two *node classification* tasks, protein function prediction and medical term semantic type classification.
- We compile 7 benchmark datasets for all the above prediction tasks and use them to systematically evaluate 11 representative graph embedding methods selected from different categories (i.e. 5 MF-based, 3 random walk-based, 3 neural network-based). We discuss our observations from extensive experiments and provide some insights and guidelines for how to choose embedding methods (including their hyper-parameter settings).
- We develop an easy-to-use Python package with detailed instructions, BioNEV (Biomedical Network Embedding Evaluation), available at: https://github.com/xiangyue9607/BioNEV, including all source code and datasets, to facilitate studying various graph embedding methods on biomedical tasks.

## 2 Overview of graph embedding methods

In this section, we provide a brief overview of different graph embedding methods that are categorized into three groups: MF-based, random walk-based and neural network-based (Fig. 1 provides a high-level illustration).

### 2.1 MF-based methods

MF has been widely adopted for data analyses. Essentially, it aims to factorize a data matrix into lower dimensional matrices and still keep the manifold structure and topological properties hidden in the original data matrix. Pioneer work in this category dates back to the early 2000s, such as Isomap (Tenenbaum *et al.*, 2000), Locally Linear Embedding (Roweis and Saul, 2000) and LEs (Belkin and Niyogi, 2003). Traditional MF has many variants, such as singular value decomposition (SVD) and graph factorization (GF) (Ahmed *et al.*, 2013). And they often focus on factorizing the first-order data matrix (e.g. adjacency matrix).

More recently, researchers focus on designing various high-order data proximity matrices to preserve the graph structure and propose various MF-based graph embedding learning methods. For example, GraRep (Cao *et al.*, 2015) considers the high-order proximity of the network and designs $k$-step transition probability matrices for factorization. HOPE (Ou *et al.*, 2016) also considers the high-order proximity. But different from GraRep, it adopts some well-known network similarity measures such as Katz Index and Common Neighbors to preserve network structures.

### 2.2 Random walk-based methods

Inspired by the word2vec (Mikolov *et al.*, 2013) model, a popular word embedding technique from Natural Language Processing (NLP), which tries to learn word representations from sentences, random walk-based methods are developed to learn node representations by generating 'node sequences' through random walks in graphs. Specifically, given a graph and a starting node, random walk-based methods first select one of the node's neighbors randomly and then move to this neighbor. This procedure is repeated to obtain node sequences. Then the word2vec model is adopted to learn embeddings based on the generated sequences of nodes. In this way, structural and topological information can be preserved into latent features.

One of the initial works in this category is DeepWalk (Perozzi *et al.*, 2014), which performs truncated random walks on a graph. Compared with DeepWalk, node2vec (Grover and Leskovec, 2016) adopts a flexible biased random walk procedure that smoothly combines breadth-first sampling and depth-first sampling to generate node sequences. Furthermore, struc2vec (Ribeiro *et al.*, 2017) is proposed for better modeling the structural identity (e.g. nodes in the network may perform similar functions). Particularly, struct2vec first constructs a multi-layer weighted graph that encodes the structural similarity between nodes where each layer $k$ is defined by using the $k$-hop neighborhoods of the nodes. DeepWalk is then performed on the multilayer graph to learn node representations in which nodes with high structural similarity are close to each other in the embedding space.

### 2.3 Neural network-based methods

Recent years have witnessed the success of neural network models in many fields. Various neural networks also have been introduced into graph embedding areas, such as multilayer perceptron (MLP) (Tang *et al.*, 2015), autoencoder (Cao *et al.*, 2016; Kipf and Welling, 2016; Wang *et al.*, 2016), generative adversarial network (GAN) (Wang *et al.*, 2018) and graph convolutional network (GCN) (Kipf and Welling, 2016, 2017). Different methods adopt different neural architectures and use different kinds of graph information as input. For example, LINE (Tang *et al.*, 2015) directly models node embedding vectors by approximating the first-order proximity and second-order proximity of nodes, which can be seen as a single-layer MLP model. DNGR (Cao *et al.*, 2016) applies the stacked denoising autoencoders on the positive pointwise mutual information (PPMI) matrix to learn deep low-dimensional node embeddings. SDNE (Wang *et al.*, 2016) adopts a deep autoencoder to preserve the second-order proximity by reconstructing the neighborhood structure of each node; meanwhile, it also incorporates LEs proximity measure into the learning framework to exploit the first-order proximity. GAE (Kipf and Welling, 2016) utilizes a GCN encoder and an inner product decoder to learn node embeddings. GraphGAN (Wang *et al.*, 2018) adopts GANs to model the connectivity of nodes. The GAN framework includes a generator and a discriminator where the generator approximates the true connectivity distribution over all other nodes and generates fake samples, while the discriminator model detects whether the sampled nodes are from ground truth or generated by the generator.

## 3 Applications of graph embedding on biomedical networks

In this section, we select 11 representative graph embedding methods (5 MF-based, 3 random walk-based, 3 neural network-based), and review how they are used on 3 popular biomedical *link prediction* applications: DDA prediction, DDI prediction, PPI prediction; and 2 biomedical *node classification* applications: protein function prediction and medical term semantic type classification.

### 3.1 Link prediction

Discovering new interactions (links) is one of the most important tasks in the biomedical area. A considerable amount of efforts has been devoted to developing computational methods to predict potential interactions in various biomedical networks, such as the DDA network (Liang *et al.*, 2017), DDI network (Zhang *et al.*, 2018b) and PPI network (Wang *et al.*, 2014). Developing such computational methods can help generate hypotheses of potential associations or interactions in biological networks.

The link prediction task can be formulated as: *given a set of biomedical entities and their known interactions, we aim to predict other potential interactions between entities* (Lü and Zhou, 2011). Traditional methods in the biomedical field put much effort on feature engineering to develop biological features [e.g. chemical substructures (Liang *et al.*, 2017), gene ontology (Gottlieb *et al.*, 2011)] or graph properties [e.g. topological similarities (Hamilton *et al.*, 2017)]. After that, supervised learning methods [e.g. support vector machine (SVM), Random Forest] (Hamilton *et al.*, 2017) or semi-supervised graph inference model [e.g. label propagation (Cowen *et al.*, 2017)] are utilized to predict potential interactions. The assumption behind these methods is that entities sharing similar biological features or graph features could have similar connections.

However, deploying methods based on biological features typically faces two problems: (i) biological features may not always be available and can be hard and costly to obtain. One popular approach to solve this problem is to remove those biological entities without features via pre-processing, which usually results in small-scale pruned datasets and thus is not pragmatic and useful in the real setting. (ii) Biological features, as well as hand-crafted graph features (e.g. node degrees), may not be precise enough to represent or characterize biomedical entities, and may fail to help build a robust and accurate model for many applications (Hamilton *et al.*, 2017).

Graph embedding methods that seek to learn node representations automatically are promising to solve the two problems mentioned above. Embedding ideas have also been employed in some recently proposed computational methods in the biomedical field. For example, MF-based techniques (Dai *et al.*, 2015; Yang *et al.*, 2014; Zhang *et al.*, 2018c) are used for predictions of DDAs. Essentially, a DDA matrix is factorized to learn low-dimensional representations for drugs and diseases in the latent space. During factorization, regularization terms or constraints can be added to further improve the quality of latent representations. For predictions of DDIs, Zhang *et al.* (2018b) propose manifold regularized MF in which Laplacian regularization is incorporated to learn a better drug representation. Besides, graph neural network is introduced for DDIs prediction (Ma *et al.*, 2018; Zitnik *et al.*, 2018) and the intuitions are similar to the GAE (Kipf and Welling, 2016). PPIs are commonly predicted using Laplacian and SVD techniques (You *et al.*, 2017; Zhu *et al.*, 2013). More recently, Wang *et al.* (2017) propose an autoencoder-based model to learn embeddings of proteins, which has a similar design to SDNE (Wang *et al.*, 2016).

### 3.2 Node classification

In addition to the link prediction task, *node classification* which aims to predict the class of unlabeled nodes given a partially labeled graph, is also one of the most important applications in graph analyses. Here, we mainly focus on two node classification applications: protein function prediction and medical term semantic type classification.

*Protein function prediction.* The large-scale experimental functional annotation of proteins is often expensive (Gligorijević *et al.*, 2018; Kulmanov *et al.*, 2018), hence graph-based computational methods which widely incorporate the idea of graph embedding, have been proposed in recent years. For example, Lim *et al.* (2018) propose a regularized Laplacian kernel-based method to learn low-dimensional embeddings of proteins. Cho *et al.* (2016) develop Mashup, which first performs random walks with restart (RWR) on PPI networks and then learns embeddings for each protein via a low rank matrix approximation method (can be optimized by SVD). The feature vectors are then fed into classifiers to derive functional insights about genes or proteins. Kulmanov *et al.* (2018) propose DeepGO that learns joint representations of proteins based on protein sequences as well as PPI network via convolutional neural nets and a graph embedding method (Alshahrani *et al.*, 2017) [similar to DeepWalk (Perozzi *et al.*, 2014)]. In node2vec, Grover and Leskovec (2016) test the effectiveness of the proposed embedding method on a PPI network. Furthermore, Zitnik and Leskovec (2017) develop OhmNet, which optimizes hierarchical dependency objectives based on node2vec to learn feature representations in multi-layer tissue networks for function prediction. Gligorijević *et al.* (2018) develop deepNF, which learns embeddings of proteins via a deep autoencoder [similar to SDNE (Wang *et al.*, 2016)].

*Medical term semantic type classification.* In the past few years, the increase of clinical texts have been encouraging data-driven models for improving the patient personal care and help clinical decision (Mullenbach *et al.*, 2018). However, due to the privacy and security concerns, the access to *raw clinical texts* is often limited (Beam *et al.*, 2018; Finlayson *et al.*, 2014; Ta *et al.*, 2018). To facilitate research on clinical texts, a popular substitute strategy for releasing raw clinical texts is to extract medical terms and their aggregated co-occurrence counts from the clinical texts (Finlayson *et al.*, 2014; Ta *et al.*, 2018). However, such released privacy-aware datasets only contain medical terms (words or phrases) extracted from clinical texts and do not reveal the semantic information (e.g. semantic types or categories). By referring to some medical knowledge bases, e.g. unified medical language system (UMLS) (Bodenreider, 2004), we can obtain semantic types (labels) medical terms. But due to mismatch and incomplete knowledge in UMLS, the semantic types of some medical terms remain unknown. Hence, we formulate a less-investigated but meaningful node classification task (Fig. 2): *given a medical term co-occurrence graph where terms and co-occurrence statistics have been extracted from clinical texts, classify the semantic types of medical terms.* In this work, we assume the clinical texts have been converted into a medical term–term co-occurrence graph as in Finlayson *et al.* (2014), where each node is an extracted medical term and each edge is the co-occurrence count of two terms in a context window. We apply graph embedding methods to the co-occurrence graph to learn representations of medical terms. Afterward, a multi-label classifier can be trained based on the learned embeddings to classify the semantic types of medical terms.

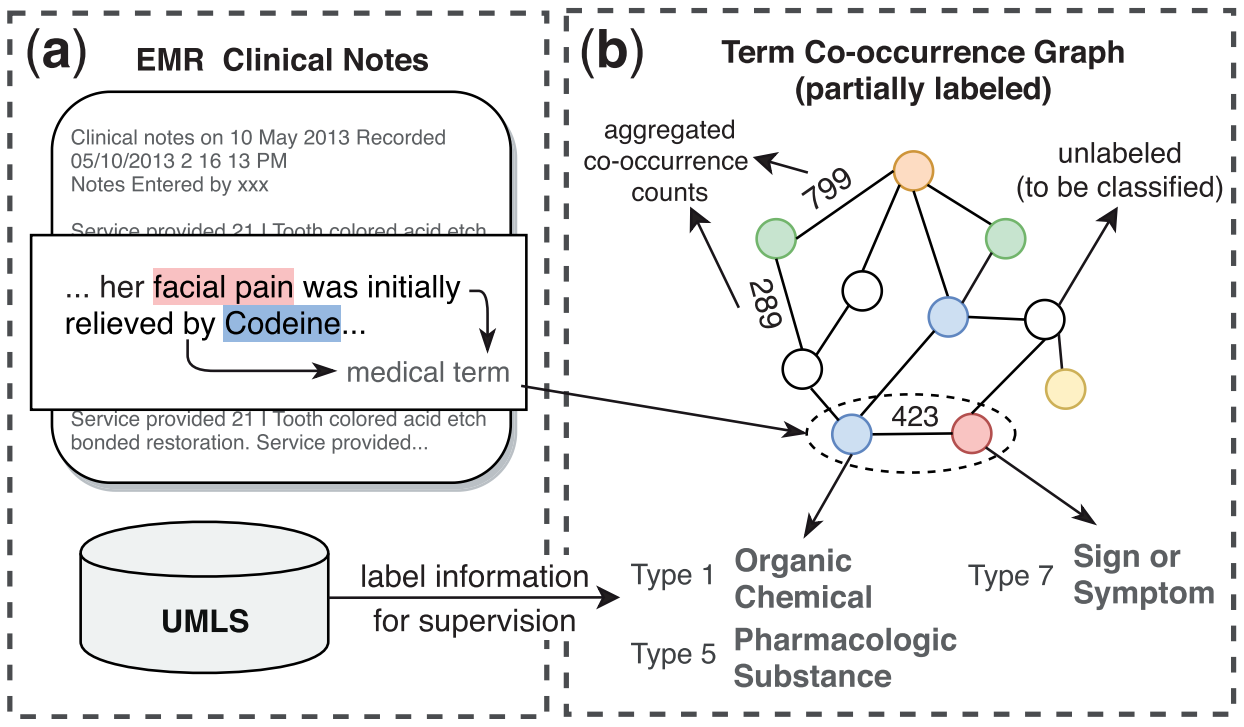

**Fig. 2.** Illustration of (**a**) how medical term–term co-occurrence graph is constructed and (**b**) node type classification in the graph. Our work assumes that the graph is given as in Finlayson *et al.* (2014) and mainly focuses on (b), i.e. testing various embedding methods on the classification performance

### 3.3 Summary

In order to show the current research status of evaluated graph embedding methods on the above biomedical applications, we summarize 11 graph embedding techniques by 3 categories and the existing works which have applied these techniques on certain tasks in Table 1. As can be seen, existing methods for the five representative biomedical applications primarily adopt the traditional techniques, e.g. LEs, MF. On the other hand, more recent advanced graph embedding methods have been demonstrated to outperform traditional techniques in social/information networks (Cao *et al.*, 2015; Grover and Leskovec, 2016; Tang *et al.*, 2015), but their performance in biomedical networks is not unknown. In addition, the comparison between these general graph embedding methods and state-of-the-arts in the individual prediction task should be explored to encourage future research. Hence, we conduct comprehensive experiments to evaluate those 11 graph embedding methods selected from 3 different categories on 5 representative biomedical tasks and compare them against the state-of-the-arts in each biomedical prediction task.

We follow the pipeline (shown in Fig. 1) of the widely adopted link prediction and node classification methods in general domains (Grover and Leskovec, 2016; Tang *et al.*, 2015): graph embeddings are first learned and then used as feature inputs to build a binary classifier or multi-label classifier (e.g. Logistic Regression, SVM, MLP) to predict the unobserved links or the node labels.

## 4 Experiments

In this section, we introduce the details of seven compiled datasets, including two DDA graphs, a DDI graph, a PPI graph for *link prediction* and a medical term–term co-occurrence graph as well as two PPI graphs for *node classification*. Then, we conduct comprehensive comparisons of 11 selected graph embedding methods on these compiled datasets.

### 4.1 Datasets

We use the following datasets for *Link Prediction*:

1. *DDA graph*. We extract chemical-disease associations from the Comparative Toxicogenomics Database (CTD) (Davis *et al.*, 2018). CTD offers two kinds of associations: curated (verified) and inferred. Since our task is to infer potential chemical-disease associations, we only use curated ones as our golden instances. Finally, we obtain 92 813 edges between 12 765 nodes (9580 chemicals and 3185 diseases) in this graph (named as 'CTD DDA').

   Also, we construct another DDA network from National Drug File Reference Terminology (NDF-RT) in UMLS (Bodenreider, 2004). NDF-RT is produced by the US Department of Veterans Affairs, and models drug characteristics including ingredients, physiologic effect and related diseases. We extract drug-disease treatment associations using the *may treat* and *may be treated by* relationships in NDF-RT. This graph (named 'NDFRT DDA') contains 13 545 nodes (12 337 drugs and 1208 diseases) and 56 515 edges.
2. *DDI graph*. We collect verified DDIs from DrugBank (Wishart *et al.*, 2018 ), a comprehensive and freely accessible online database that contains detailed information about drugs and drug

**Table 1.** A summary of 11 representative graph embedding methods and existing work (if any) using them for a certain task

| Method category | | Method name | Link prediction tasks | | | Node classification tasks | |
|---|---|---|---|---|---|---|---|
| | | | Drug-disease association prediction | Drug–drug interaction prediction | Protein–protein interaction prediction | Medical term type classification | Protein function prediction |
| Traditional | Matrix factorization-based | Laplacian (Belkin and Niyogi, 2003) | Zhang *et al.* (2018c) | (Zhang *et al.*, 2018b) | Zhu *et al.* (2013) | ✗ | Lim *et al.* (2018) |
| | | SVD | Dai *et al.* (2015) | ✗ | You *et al.* (2017) | ✗ | Cho *et al.* (2016) |
| | | GF (Ahmed *et al.*, 2013) | Yang *et al.* (2014) and Zhang *et al.* (2018c) | (Zhang *et al.*, 2018b) | ✗ | ✗ | ✗ |
| Recently Proposed | | HOPE (Ou *et al.*, 2016) | ✗ | ✗ | ✗ | ✗ | ✗ |
| | | GraRep (Cao *et al.*, 2015) | ✗ | ✗ | ✗ | ✗ | ✗ |
| | Random walk-based | DeepWalk (Perozzi *et al.*, 2014) | ✗ | ✗ | ✗ | ✗ | Cho *et al.* (2016) and Kulmanov *et al.* (2018) |
| | | node2vec (Grover and Leskovec, 2016) | ✗ | ✗ | ✗ | ✗ | Grover and Leskovec (2016) and Zitnik and Leskovec (2017) |
| | | struc2vec (Ribeiro *et al.*, 2017) | ✗ | ✗ | ✗ | ✗ | ✗ |
| | Neural network-based | LINE (Tang *et al.*, 2015) | ✗ | ✗ | ✗ | ✗ | ✗ |
| | | SDNE (Wang *et al.*, 2016) | ✗ | ✗ | Wang *et al.* (2017) | ✗ | Gligorijević *et al.* (2018) |
| | | GAE (Kipf and Welling, 2016) | ✗ | Zitnik *et al.* (2018) and Ma *et al.* (2018) | ✗ | ✗ | ✗ |

*Note*: ✗ means that a method (row) has not been applied for a task (column).

targets. We obtain 242 027 DDIs between 2191 drugs and refer to this dataset as 'DrugBank DDI'.

3. *PPI graph*. We extract *Homo sapiens* PPIs from STRING database (Szklarczyk *et al.*, 2015). Each PPI is associated with a confidence score that indicates its possibility to be a true positive interaction. To reduce noise, we only collect PPI whose confidence score is larger than 0.7 according to the guidelines of STRING database. Finally, we obtain 359 776 interactions among 15 131 proteins and name this dataset as 'STRING PPI'.

We use the following datasets for *Node Classification*:

1. *Medical term–term co-occurrence graph*. We adopt a publicly available set of medical terms with their co-occurrence statistics which are extracted by Finlayson *et al.* (2014) from 20 million clinical notes collected from Stanford Hospitals and Clinics (Lowe *et al.*, 2009) since 1995. Medical terms are extracted from raw clinical notes using an existing phrase mining tool (LePendu *et al.*, 2012) by matching with 22 clinically relevant ontologies such as SNOMED-CT and MedDRA. Co-occurrence frequencies between two terms are counted based on how many times they co-occur in the same temporal *bin* (i.e. a certain timeframe; see Finlayson *et al.*, 2014 for more details). We select *perBin 1-day* dataset since it contains more medical terms compared with other bins. To filter very common medical terms (e.g. 'medical history', 'medication dose') that may influence the quality of embeddings, we convert the co-occurrence counts to the PPMI value (Levy and Goldberg, 2014) and remove the edges whose PPMI value is <2. We also adopt a subsampling (Mikolov *et al.*, 2013) strategy to further filter common terms and construct a medical term–term co-occurrence graph that contains 48 651 medical terms and 1 659 249 edges.
We keep the medical terms that can be mapped to the UMLS Concept Unique Identifiers (CUI) and collect their corresponding semantic types (e.g. clinical drug, disease or syndrome) from UMLS. We select 31 different semantic types, with each having more than 20 samples. Finally, we obtain 25 120 nodes with label information. This dataset is called 'Clin Term COOC'.
2. *PPI graphs with functional annotations*. We also compile two PPI graphs with functional annotations from previous studies. One is from node2vec (Grover and Leskovec, 2016), which contains 3890 proteins, 76 584 interactions and 50 different function annotations (labels). This dataset is named as 'node2vec PPI'. The other one is from Mashup (Cho *et al.*, 2016), which is designed for integrating different information from multiple networks. The Mashup dataset contains six individual PPI networks (e.g. experimental, coexpression). Given that our selected graph embedding methods can only work on a single network, we select the *experimental* PPI network (there are six individual PPI networks in Mashup dataset, we select the *experimental* PPI network since Mashup achieves the best performance on it under single-network circumstance) to learn embeddings. The *experimental* PPI network contains 300 181 interactions between 16 143 proteins. Same to Mashup, we use the 3 grouped distinct levels of functional categories of varying specificity, each containing 28 100 and 262 different annotations, respectively. We only adopt the first level (28 labels) for the main comparison experiment for simplicity. Other label information is used in comparing the recent embedding methods with Mashup in Section 4.4. This dataset is called 'Mashup PPI'.

The details of all datasets are summarized in Table 2.

## 4.2 Experimental set-up

We use OpenNE (https://github.com/thunlp/OpenNE), an open-source Python package for network embedding, to learn node embeddings for LEs (Belkin and Niyogi, 2003), HOPE (Ou *et al.*, 2016), GF (Ahmed *et al.*, 2013), DeepWalk (Perozzi *et al.*, 2014), LINE (Tang *et al.*, 2015) and SDNE (Wang *et al.*, 2016). We run SVD using Numpy (http://www.numpy.org/) and obtain struc2vec (https://github.com/leoribeiro/struc2vec) (Ribeiro *et al.*, 2017) and GAE (https://github.com/tkipf/gae) (Kipf and Welling, 2016) embeddings using the source code provided by their authors. More implementation details can be found in Supplementary Material.

**Table 2.** Statistics of the datasets, where the *Density* is defined as $2 \times$ no. edges/no. nodes$^2$

| Task type | Dataset | No. nodes | No. edges | Density | No. node labels |
|---|---|---|---|---|---|
| Link prediction | CTD DDA | 12 765 | 92 813 | 0.11% | — |
| | NDFRT DDA | 13 545 | 56 515 | 0.06% | — |
| | DrugBank DDI | 2191 | 242 027 | 10.08% | — |
| | STRING PPI | 15 131 | 359 776 | 0.31% | — |
| Node classification | Clin Term COOC | 48 651 | 1 659 249 | 0.14% | 31 |
| | node2vec PPI | 3890 | 76 584 | 1.01% | 50 |
| | MashUp PPI | 16 143 | 300 181 | 0.23% | 28 |

For the link prediction tasks (Section 4.3), all the known interactions are positive samples and are split into the training set (80%) and testing set (20%). Since unknown interactions are far more than known ones, we randomly select disconnected edges as negative samples with an equal number of positive samples in both training and testing phase. For each node pair, we concatenate the embeddings of two nodes as the edge feature and then build a Logistic Regression binary classifier based on it using scikit-learn package (Pedregosa *et al.*, 2011). Area under ROC curve (*AUC*), *accuracy* and *F1* score are used to evaluate the performance of the classifiers, so as to evaluate different embedding methods.

For the node classification task (Section 4.4), we use the entire graph information to train the embeddings. Nodes with label information are then split into the training set (80%) and the testing set (20%). The embedding vectors of nodes are directly treated as feature vectors and used to train *One-vs-Rest* Logistic Regression classifiers using the scikit-learn package. We assign top $\alpha_i$ predictions to the node $i$ as its predicted labels, where $\alpha_i$ is the number of golden labels of the node $i$ in the testing set. *Accuracy*, *Macro-F1* and *Micro-F1* are used to evaluate the performance of different embedding methods on the testing set. Accuracy is defined as the percentage of samples that have all their labels classified correctly. *F1* score is the harmonic mean of precision and recall. We adopt two weighted strategies of *F1* score: micro (calculate metrics globally by counting the total true positives, false negatives and false positives) and macro (calculate metrics for each label, and find their unweighted mean).

For all embedding methods, the dimensionality of the learned embedding is set to 100 unless otherwise stated (we also discuss its impact on the performance in Section 4.5). Moreover, we tune 1–2 significant hyper-parameters for some embedding methods via grid-search (see Section 4.5 for details). Other hyper-parameters for each method are set at their default values recommended by the corresponding papers.

### 4.3 Link prediction results

We conduct the link prediction task on the 4 compiled biomedical networks: CTD DDA, NDFRT DDA, DrugBank DDI and STRING PPI. Table 3 shows the overall performance of different embedding methods on the four datasets.

Generally, compared with traditional techniques (e.g. LEs, SVD, GF), the recently proposed embedding methods have largely improved the link prediction performance. For example, LINE achieves 3–23% improvement in terms of *AUC* value on the four datasets compared with LEs. Struc2vec obtains 3–15% increment in the *accuracy* on the four datasets, respectively, when compared with GF. These results demonstrate that the recently proposed graph embedding methods are more effective and could be used on various biological link prediction tasks to improve the prediction performance.

Furthermore, we have the following key observations and analyses:

- *For the MF-based methods*, since HOPE and GraRep are designed to capture the high-order proximity of graphs, they are usually more effective than traditional MF methods that only preserve the first-order of networks.
- *For the random walk-based methods*, generally, struc2vec performs better than DeepWalk and node2vec. This is because compared with DeepWalk and node2vec, struc2vec constructs a hierarchy weighted graph to measure the structural identity. Such hierarchy structure design incorporates both node degree distributions from the bottom as well as the entire network on the top, which can better capture the graph structure information and hence obtain better performance.
- *For the neural network-based methods*, LINE achieves competitive prediction performance consistently when compared with the best performing method on each dataset. It indicates that directly modeling edge information by a single-layer MLP is an effective way to learn node embeddings. SDNE and GAE also obtain satisfying prediction performance, which demonstrates that autoencoders and GCNs can also be useful for capturing graph structural information.

*Comparison with state-of-the-art studies.* To further demonstrate the effectiveness of graph embedding methods, we compare them with the state-of-the-art methods for two link prediction: DDA prediction and DDI prediction.

For the DDAs prediction, we select LRSSL (Liang *et al.*, 2017) as our baseline. LRSSL is a Laplacian regularized sparse subspace learning framework which aims to project different drug features into a common subspace. Three drug feature profiles (i.e. chemical substructure, target domain and target annotation) are used in the training process. To be fair, we adopt the code and dataset used in the LRSSL. To learn graph embeddings without modeling biological features, we run four representative graph embedding methods: GraRep, DeepWalk, LINE and struc2vec on LRSSL's DDA graph. Following the same train/test split, training and evaluation process of link prediction in Section 4.2, we plot the ROC Curves to illustrate the performance of different methods better. As seen in Figure 3a, graph embedding methods achieve competitive performance compared with LRSSL. Furthermore, we use the learned DeepWalk embedding vectors as the fourth feature for the LRSSL method and improve the LRSSL performance, which indicates that the learned node embedding can be used as a *complementary representation* for biological features.

For the DDIs prediction, we compare the embedding methods with a recent method DeepDDI (Ryu *et al.*, 2018). DeepDDI first adopts principal component analysis to reduce the dimension of the drug features (i.e. drug substructure) and then feeds these into a deep neural network (DNN) classifier. For a fair comparison with graph embedding methods and to reduce the bias caused by different classifiers, we compare these methods under four classifiers, Naive Bayes, Linear SVM, Logistic Regression and eight-layer DNN (the same as the original paper). More implement details can be found in Supplementary Material. As seen in Figure 3b, graph embedding methods outperform the drug features-based model or obtain very competitive performance under each classifier, which demonstrates the power of graph embedding methods.

### 4.4 Node classification results

Table 4 shows the performance of different embedding methods on medical term semantic type classification and protein function prediction. We make the following key observations:

- *For the MF-based methods*, it is a little surprising that the traditional method SVD achieves better performance, even surpassing HOPE and GraRep. This may indicate that directly modeling

**Table 3.** Overall link prediction performance on the four compiled biomedical datasets

| Method category | | Method name | CTD DDA | NDFRT DDA | DrugBank DDI | STRING PPI |
|---|---|---|---|---|---|---|
| Traditional | Matrix factorization-based | Laplacian (Belkin and Niyogi, 2003) | 0.856±0.004 | 0.930±0.003 | 0.796±0.002 | 0.639±0.021 |
| | | SVD | 0.936±0.002 | 0.779±0.003 | 0.919±0.001 | 0.867±0.001 |
| | | GF (Ahmed *et al.*, 2013) | 0.884±0.004 | 0.720±0.006 | 0.882±0.003 | 0.817±0.005 |
| Recently proposed | | HOPE (Ou *et al.*, 2016) | 0.951±0.001 | 0.949±0.001 | 0.923±0.001 | 0.839±0.001 |
| | | GraRep (Cao *et al.*, 2015) | **0.960±0.001** | **0.963±0.001** | **0.925±0.001** | **0.894±0.001** |
| | Random walk-based | DeepWalk (Perozzi *et al.*, 2014) | 0.929±0.002 | 0.783±0.004 | **0.921±0.001** | 0.884±0.001 |
| | | node2vec (Grover and Leskovec, 2016) | 0.911±0.002 | 0.819±0.005 | 0.902±0.001 | 0.828±0.003 |
| | | struc2vec (Ribeiro *et al.*, 2017) | **0.965±0.001** | **0.958±0.001** | 0.904±0.001 | **0.909±0.001** |
| | Neural network-based | LINE (Tang *et al.*, 2015) | **0.965±0.001** | **0.962±0.002** | 0.905±0.002 | 0.859±0.003 |
| | | SDNE (Wang *et al.*, 2016) | 0.935±0.010 | 0.944±0.004 | 0.911±0.006 | 0.884±0.008 |
| | | GAE (Kipf and Welling, 2016) | 0.937±0.001 | 0.813±0.007 | **0.917±0.001** | **0.900±0.001** |

*Note*: Due to the limited space, we only show the *AUC* value. Other evaluation metrics can be found in Supplementary Material. The best performing method in each category is in bold.

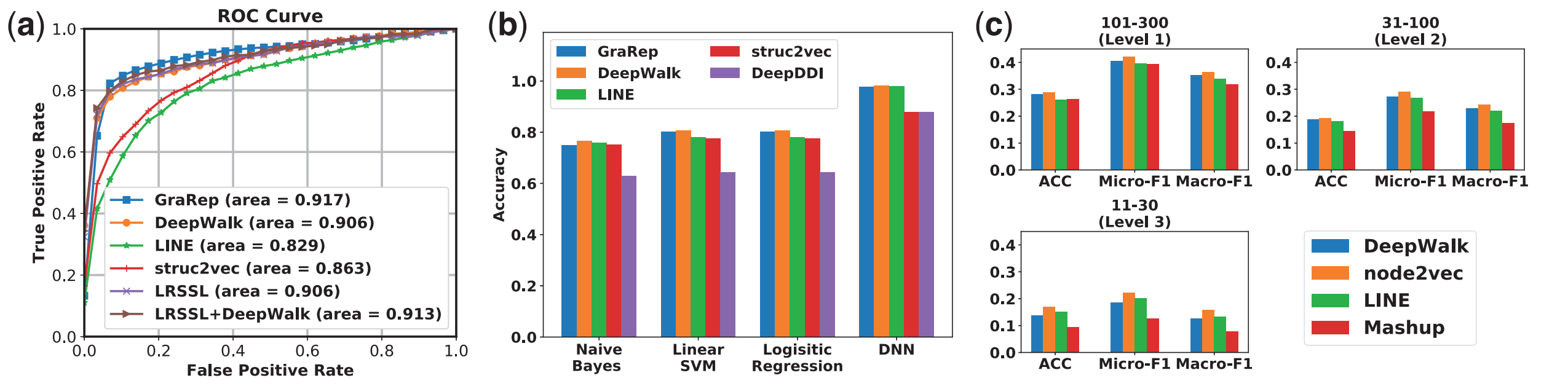

**Fig. 3.** (**a**) Comparison with the state-of-the-arts for drug-disease association prediction (LRSSL) (Liang *et al.*, 2017); (**b**) drug–drug interaction prediction (DeepDDI) (Ryu *et al.*, 2018) and (**c**) gene (protein) function prediction (Mashup) (Cho *et al.*, 2016). Same as Mashup, we evaluate their performance on three-level human Biological Process (BP) gene annotations (each containing GO terms with 101–300, 31–100 and 11–30 genes, respectively). As can be seen, in each task, general graph embedding methods achieve competitive performance against them

the first-order proximity would be good enough to classify the nodes.

- *For the random walk-based methods*, node2vec performs better since it aims to capture different functions of nodes (i.e. homophily and structural equivalence) via a more flexible biased random walk. Struc2vec performs not good as DeepWalk and node2vec as it mainly focuses on modeling the structural identity of nodes; however, a clear structural role may not exist in these biomedical graphs and struc2vec is not suitable on such graphs.
- *For the neural network-based methods*, LINE achieves better performance than SDNE, which demonstrates that directly modeling edge information is an effective way to learn the embedding for the node classification task. And GAE also achieve promising performance, which demonstrates the power of the graph neural networks.

*Comparison with state-of-the-art study.* To better illustrate the effectiveness of the recent graph embedding methods in biomedical node classification tasks, we select protein function prediction as our representative node classification task and compare the graph embedding methods with a popular state-of-the-art: Mashup (Cho *et al.*, 2016).

Mashup is also one of embedding learning methods. But different from other embedding methods which learn node embedding in a single network, Mashup is carefully designed to diffuse the information from multi-networks. Specifically, RWR is firstly used to compute the diffusion state for each node in each individual network. Low-dimensional embeddings are then obtained by jointly minimizing the difference between the observed diffusion states and the parameterized-multinomial logistic distributions across all networks. To make a fair comparison with Mashup, we construct a diffusion PPI network by doing a simple unweighted sum of each interaction score in the individual networks and then run different embedding methods on this simple diffusion network. As seen in Figure 3c, the three representative graph embedding methods: DeepWalk, node2vec and LINE achieve very competitive or better performance compared with Mashup on three-level protein function prediction.

Mashup is specially designed for protein/gene-related prediction tasks and has an advanced network diffusion strategy (e.g. jointly optimizing the embedding based on information from each individual network), but the recent embedding methods can still achieve competitive performance. This may give some inspirations for future study (e.g. considering to replace the current embedding optimization process of Mashup with DeepWalk, node2vec or LINE).

## 4.5 Influence of hyper-parameters

Hyper-parameters play essential roles in machine learning models. However, selecting proper hyper-parameters is often time-consuming. We investigate the influence of some important hyper-parameters in various embedding methods. By running grid-search of these important hyper-parameters of each method, we expect to summarize some general guidelines for helping researchers better set the hyper-parameters, so as to save their time and efforts.

We first evaluate how different embedding dimensions can affect the prediction performance and time efficiency. Figure 4 shows the impact of embedding dimensionality on the prediction performance and time efficiency for ‘CTD DDA’ dataset. Generally, the prediction performance becomes better when the embedding dimensionality increases, which is intuitive since higher dimensionality can

**Table 4.** Overall node classification performance on the three compiled datasets

| Method category | Method name | Clini COOC | | node2vec PPI | | Mashup PPI | |
|---|---|---|---|---|---|---|---|
| | | Micro-F1 | Macro-F1 | Micro-F1 | Macro-F1 | Micro-F1 | Macro-F1 |
| Matrix factorization-based | Laplacian (Belkin and Niyogi, 2003) | 0.313±0.005 | 0.073±0.002 | 0.101±0.008 | 0.070±0.007 | 0.132±0.009 | 0.107±0.008 |
| | SVD | 0.420±0.005 | 0.186±0.007 | 0.228±0.011 | 0.179±0.011 | 0.347±0.014 | 0.297±0.014 |
| | GF (Ahmed *et al.*, 2013) | 0.352±0.007 | 0.143±0.009 | 0.168±0.011 | 0.121±0.011 | 0.290±0.015 | 0.237±0.016 |
| | HOPE (Ou *et al.*, 2016) | 0.395±0.005 | 0.163±0.006 | 0.208±0.011 | 0.152±0.011 | 0.322±0.013 | 0.266±0.013 |
| | GraRep (Cao *et al.*, 2015) | **0.424±0.006** | **0.177±0.005** | **0.238±0.010** | **0.193±0.013** | **0.334±0.011** | **0.283±0.011** |
| Random walk-based | DeepWalk (Perozzi *et al.*, 2014) | 0.472±0.005 | 0.227±0.007 | **0.243±0.001** | **0.194±0.011** | 0.357±0.011 | 0.311±0.012 |
| | node2vec (Grover and Leskovec, 2016) | **0.479±0.005** | **0.231±0.010** | **0.243±0.009** | 0.190±0.011 | **0.367±0.012** | **0.313±0.013** |
| | struc2vec (Ribeiro *et al.*, 2017) | 0.253±0.006 | 0.038±0.001 | 0.094±0.006 | 0.061±0.004 | 0.120±0.010 | 0.087±0.008 |
| Neural network-based | LINE (Tang *et al.*, 2015) | **0.453±0.006** | **0.205±0.008** | 0.236±0.011 | 0.176±0.012 | 0.352±0.017 | 0.296±0.017 |
| | SDNE (Wang *et al.*, 2016) | 0.271±0.016 | 0.042±0.007 | 0.098±0.010 | 0.047±0.007 | 0.178±0.013 | 0.109±0.012 |
| | GAE (Kipf and Welling, 2016)[a] | — | — | **0.237±0.014** | **0.186±0.014** | **0.358±0.013** | **0.307±0.014** |

*Note*: The best performing method in each category is in bold.

[a]The source code of GAE provided by the authors does not support a large-scale graph (nodes>40k). We omit its performance on 'Clini COOC' here.

encode more useful information. Then, the performance tends to saturate when the dimension reaches to a threshold (e.g. 100). As for the time cost, it first increases gradually below 100 but tends to boost sharply (the *y*-axis is log-based) if the dimensionality continues to increase. So we would not suggest to set the dimensionality to be too large (e.g. around 100 is a good option) for the practitioners when considering both performance and time efficiency. The results of dimensionality's influence on other datasets can be found in Supplementary Figures S1 and S2.

Furthermore, we choose sensitive hyper-parameters for 7 embedding methods, which have been pointed out to be important by their authors. Table 5 shows the selected hyper-parameters in different embedding methods as well as their meanings. We spend a lot of efforts on carefully tuning these hyper-parameters by grid search. The influence of the hyper-parameters on each embedding method is shown in Supplementary Figures S3–S9, respectively. By summarizing these results, we provide some high-level guidelines on setting hyper-parameters for practitioners in Table 5.

## 4.6 Summary of experimental results

To better help the practitioners select proper embedding methods for their biomedical prediction task, we summarize the experimental results and discuss our observations:

- Generally, the recently proposed graph embedding methods achieve very promising results in various biomedical prediction tasks. They deserve more attention for future biomedical graph analysis.
- By simply applying the recent graph embedding methods on biomedical graphs and then feeding into a classifier, we can achieve very competitive or better performance compared with state-of-the-arts. Future model design for biomedical prediction tasks may begin at these embedding methods or integrate them as one module into the proposed method, which is expected to gain better results.
- In particular, for MF-based methods, we observe that modeling high-order proximity (e.g. HOPE, GraRep) is generally useful for link prediction tasks on medical graphs but may be less meaningful for the node classification tasks. For random walk-based methods, struc2vec is more suitable for link prediction tasks (when there is a lack of structural identity in graphs) while node2vec and DeepWalk are more suitable for node classification tasks. For neural network-based methods, LINE usually achieves competitive performance against the best performing method on each dataset. SDNE can achieve good performance on link prediction tasks but less satisfying performance on node classification. GAE performs well in relatively large-scale network but may not perform well on small-scale datasets.

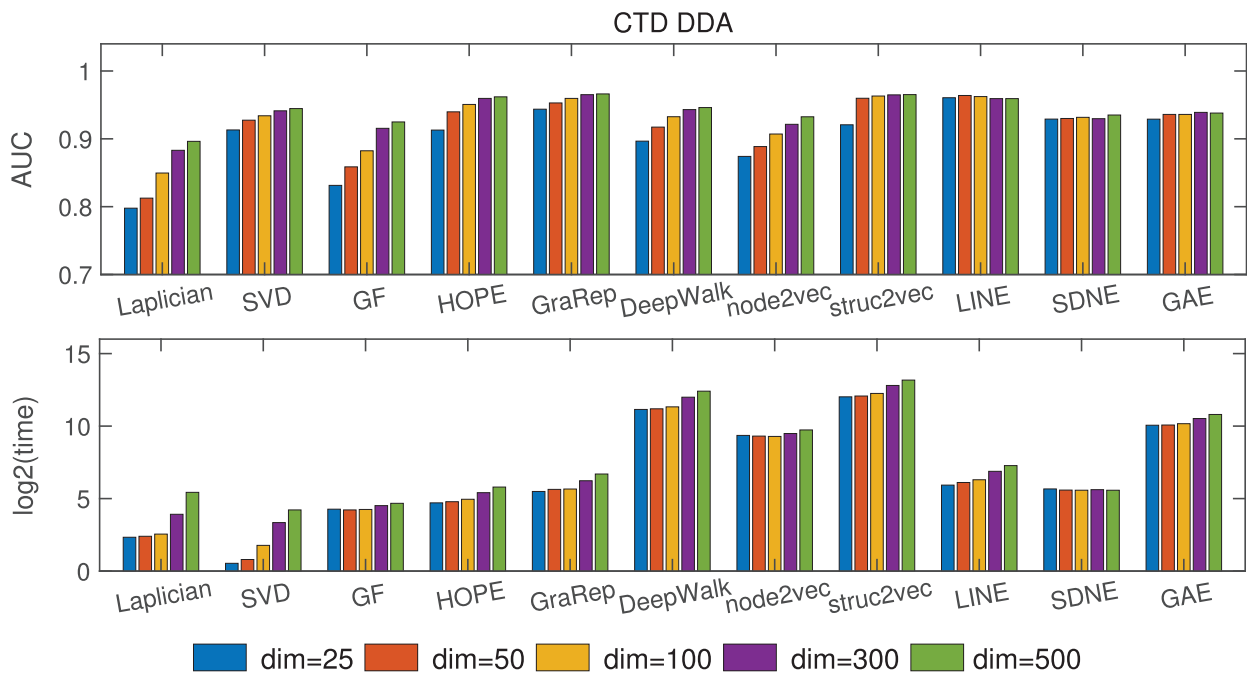

**Fig. 4.** The influence of dimensionality on the performance and training time of different embedding methods based on 'CTD DDA' dataset

More details of the datasets, implementation, experiment results, guidelines can be found in Supplementary Material.

## 5 Discussions and future directions

*Connections of network embedding and network propagation.* In the recent biomedical network analyses, a very popular paradigm is network propagation (Cowen *et al.*, 2017), which amplifies a biological signal (e.g. label, association) based on the assumption that nodes with similar neighbors (e.g. genes underlying similar phenotypes) tend to interact with one another (Menche *et al.*, 2015). Specifically, the information of one node is propagated through the edges to their neighbors in an iterative manner for a fixed number of steps or until convergence (Cowen *et al.*, 2017). The core of these propagation methods is random walk, which is also adopted in many embedding methods (e.g. Deepwalk, node2vec and struc2vec). But different from network diffusion, which propagates the 'signal' in the network directly, the random walk-based embedding methods treat the 'walk' as a kind of node similarity or proximity characterizing method. They expect to preserve the network structural information as much as possible through a fixed number of random walks. These 'walking histories' (i.e. node sequences) are then fed into word2vec (Mikolov *et al.*, 2013) to learn low-dimensional embeddings. Though the pipeline of the random walk-based embedding methods and network propagation methods is different, their idea and assumption are similar. They both assume that nodes with similar neighbors have similar functions and tend to interact with each other. Besides random walk-based embedding methods, this assumption is also widely adopted in other embedding methods (e.g. LINE, SDNE).

**Table 5.** Meanings of main hyper-parameters in different embedding methods and general guidelines for setting hyper-parameters of these embedding methods

| Methods | Hyper-parameters | General guidelines |
|---|---|---|
| GraRep (Cao *et al.*, 2015) | *Ksteps*: $k$-step relational information ($k$-step transition probability matrix) | A large value for link prediction tasks (e.g. 3, 4); a small value for node classification tasks (e.g. 1, 2) |
| DeepWalk (Perozzi *et al.*, 2014) | *Number of walks*: the number of walks at each node; *walk length*: the length of each walk; | Large values for both (e.g. 64 128 256) |
| node2vec (Grover and Leskovec, 2016) | $p$, $q$: two parameters that control how fast the walk explores and leaves the neighborhood of starting node | Vary from graphs to graphs, may tune at small values for both (e.g. 0.25) |
| struc2vec (Ribeiro *et al.*, 2017) | *Number of walks*: the number of walks at each node; *walk length*: the length of each walk | Large values for both (e.g. 64 128 256) |
| LINE (Tang *et al.*, 2015) | *epochs*: number of training epochs | Small training epochs for small-scale graphs (e.g. 5); and large value for large-scale graph (e.g. 20) |
| SDNE (Wang *et al.*, 2016) | $\alpha$: balances the weight of first-order and second-order proximities; $\beta$: controls the reconstruction weight of the non-zero elements in the training graph | Vary from graphs to graphs, may tune at small values for both (e.g. $a$=0.1, $b$=0) |
| GAE (Kipf and Welling, 2016) | *Hidden units*: number of units in hidden layer | A large value (e.g. 128) |

Additionally, there are some variants of random walk, e.g. random walk with restart (RWR), personalized PageRank and diffusion kernel. They also involve the embedding ideas, e.g. using Laplacian normalized matrix, factorizing inverse Laplacian matrix. These variants can also be incorporated into current random walk-based embedding framework.

*Modeling external information in graphs.* In addition to the graph structure, external information can also help build computational models for biomedical networks. For example, Zhang *et al.* (2018c) incorporate drug and disease features into MF to learn better representations. Žitnik and Zupan (2014) incorporate prior information (e.g. gene network) as a vector or a matrix to further improve the gene-related prediction tasks. There may also exist partial label information on graphs (e.g. semantic types are partly available for nodes in a medical term co-occurrence graph). Incorporating those features and labels into advanced graph embedding models can potentially further improve the performance. There have been a surge of *attributed graph embedding* methods that explore this direction. For example, DDRW (Li *et al.*, 2016) and MMDW (Tu *et al.*, 2016) jointly optimize the objective of DeepWalk with an SVM classification loss to incorporate label information. We leave benchmarking such *attributed network embedding* methods on biomedical graphs as our future work.

*Transfer learning for graph embedding.* Recent studies in Computer Vision and NLP show that *transfer learning* helps improve model performance on different tasks (Howard and Ruder, 2018; Shin *et al.*, 2016). General patterns are captured during pre-trained processes and can be 'transferred' into new prediction tasks. There also exist some pre-trained embeddings of biomedical entities (Beam *et al.*, 2018; Choi *et al.*, 2016) which allow us to adopt similar ideas of 'transfer learning' to learn graph embeddings. We can initialize the embedding vector for each node on a graph with its pre-trained embedding (e.g. by looking for the corresponding entity in Choi *et al.*, 2016; Beam *et al.*, 2018) rather than by random initialization, and then continue training various graph embedding methods as before (which is often referred to as 'fine-tuning'). The pre-trained embeddings can be seen as 'coarse embeddings' since they are usually pre-trained on a large general corpus and have not been optimized for downstream tasks yet. Nevertheless, they contain some additional semantic information that may not be able to be learned from a downstream task graph (e.g. due to its small scale). By fine-tuning, such additional semantic information can be 'transferred' into the finally learned embeddings. We conduct experiment with this transfer learning idea on the 'CTD DDA' graph. As seen in Supplementary Table S3, the link prediction performance has been improved using the pre-trained embeddings from Beam *et al.* (2018). Currently, the number of released biomedical entities with pre-trained embeddings is still limited and entities without pre-trained embeddings have to be initialized randomly. However, with the increasing volume of biomedical data, more and more entities can have pre-trained embeddings, and the idea of *pre-training—then—fine-tuning* can be more promising.

## 6 Conclusion

This paper provides an overview of various graph embedding techniques and evaluates their performance on two important biomedical tasks, link prediction and node classification. Specifically, we compile 7 datasets from public database or previous studies and use them to benchmark 11 representative graph embedding methods. Through extensive experiments, we find that generally the recent graph embedding methods can perform well in various biomedical prediction tasks and can also achieve very competitive or better performance compared with state-of-the-arts. Hence, these recent graph embedding methods can be considered as a starting point when designing advanced models for future biomedical prediction tasks. Additionally, we tune some important hyper-parameters of graph embedding methods and provide general guidelines for setting hyper-parameters for practitioners. We also discuss the connections between the recent network propagation (diffusion) methods and the graph embedding methods as well as potential directions (e.g. transfer learning for graph embedding) to inspire the future work.

## Acknowledgements

We thank Dr Deborah A. Petrone, Kaushik Mani and anonymous reviewers for their helpful comments and suggestions on our work, and Ohio Supercomputer Center (OSC) (Ohio Supercomputer Center, 1987) for providing us computing resources.

## Funding

This work has been supported by Patient-Centered Outcomes Research Institute (PCORI) under grant ME-2017C1-6413.

*Conflict of Interest*: none declared.